\documentclass[letterpaper, 10 pt, conference]{ieeeconf}  

\usepackage{graphics}
\usepackage{graphicx} 
\usepackage{rotating}
\usepackage{longtable}
\usepackage{multirow}
\usepackage{amsmath}
\usepackage{amssymb}
\usepackage{xspace}
\usepackage{textcase}
\usepackage{subcaption}
\usepackage{dblfloatfix}
\usepackage{booktabs}
\usepackage{float}

\usepackage{mathtools}
\usepackage{hyperref}

\DeclareMathOperator*{\argmin}{arg\,min}

\usepackage{algorithm}
\usepackage{algpseudocode}

\IEEEoverridecommandlockouts                              

\title{\LARGE \bf
Metric Learning to Accelerate Convergence of Operator Splitting Methods for Differentiable Parametric Programming
}

\author{Ethan King, James Kotary, Ferdinando Fioretto, J\'an Drgo\v na
\thanks{Ethan King is with the Pacific Northwest National Laboratory, Richland, WA, USA
        {\tt\small ethan.king@pnnl.gov}}%
\thanks{James Kotary is with the University of Virginia, Charlottesville, Virginia, USA
        {\tt\small jk4pn@virginia.edu}}%
\thanks{Ferdinando Fioretto is with the University of Virginia, Charlottesville, Virginia, USA
        {\tt\small fioretto@virginia.edu}}%
\thanks{J\'an Drgo\v na is with the Pacific Northwest National Laboratory, Richland, WA, USA
        {\tt\small Jan.Drgona@pnnl.gov}}%
}

\begin{document}

\maketitle
\thispagestyle{empty}
\pagestyle{empty}

\begin{abstract}

Recent work has shown a variety of ways in which machine learning can be used to accelerate the solution of constrained optimization problems. Increasing demand for real-time decision-making capabilities in applications such as artificial intelligence and optimal control has led to a variety of approaches, based on distinct strategies.  This work proposes a novel approach to learning optimization, in which the underlying metric space of a proximal operator splitting algorithm is learned so as to maximize its convergence rate. While prior works in optimization theory have derived optimal metrics for limited classes of problems, the results do not extend to many practical problem forms including general Quadratic Programming (QP). This paper shows how differentiable optimization can enable the end-to-end learning of proximal metrics, enhancing the convergence of proximal algorithms for QP problems beyond what is possible based on known theory. Additionally, the results illustrate a strong connection between the learned proximal metrics and active constraints at the optima, leading to an interpretation in which the learning of proximal metrics can be viewed as a form of active set learning.

\end{abstract}

\section{INTRODUCTION}

A substantial literature has been dedicated to the use of machine learning to aid in the fast solution of constrained optimization problems. This interest is driven by an increasing need for real-time decision-making capabilities, in which decision processes modeled by optimization problems must be resolved faster than can be met by traditional optimization methods. Such capabilities are of interest in various application settings such as job scheduling in manufacturing \cite{kotary2022fast,kotary2021learning}, power grid operation \cite{fioretto2020lagrangian}, and optimal control \cite{sambharya2023end}.

A prominent application of machine learning in accelerating optimization is to learn the parameters of a standard solution algorithm, such that iterations to convergence are minimized. Examples include gradient stepsizes \cite{amos2023tutorial}, and initial solution estimates \cite{sambharya2023end}.  This paper proposes an alternative approach by parametrizing the underlying metric space of an optimization algorithm which relies on \emph{proximal operators}. Proximal operators, which include projections, are based on a notion of distance within a metric space and employed in many practical optimization methods. While most methods that employ proximal algorithms are typically based on the standard Euclidean metric, it is well-known that many such methods are also guaranteed to converge for non-Euclidean metrics defined as general quadratic forms over the continuous space of positive definite matrices \cite{beck2017first}. 

The possibility of accelerating convergence by selecting non-Euclidean metrics within that space has been noted \cite{giselsson2015metric}, but no known method has shown to be effective over a general class of optimization problems. For limited classes of problems, \emph{optimal} metric choices have been modeled as the solution to an auxiliary optimization problem. But for many problems including general quadratic programming (QP) problems, such models are yet unknown \cite{Giselsson17}. Theoretical insights have been used to suggest \emph{heuristic} metric choices for QP problems \cite{giselsson2015metric,Giselsson17}, but the potential for improvement over these heuristic rules has not been fully explored.


This paper proposes differentiable programming to both explore the potential, and to overcome the challenges of metric selection for more general classes of optimization problems. Specifically, we propose a system of end-to-end learning for proximal optimization, which trains machine learning models to predict metrics that empirically minimize solution error over a prescribed number of iterations on a given problem instance. 
Enhanced convergence of two proximal optimization methods is demonstrated on Quadratic Programming (QP) problems, including test cases where theoretically prescribed heuristic metric choices perform poorly. 

We demonstrate that while prior heuristic models of optimal metric selection can \emph{fail} in the presence of active constraints at the optimal solution, our learned  metrics are \emph{correlated} with the active constraints at optima, and can accelerate convergence by ignoring the inactive constraints. 
This leads to an interpretation of metric selection as a problem which incorporates active set prediction, whose difficulty may approach that of solving the optimization problem itself. The proposed integration optimization and learning thus shows advantages in both accuracy and efficiency over theoretical approaches to metric selection in proximal optimization.


\section{Related Work}
This paper's topic is at the intersection of learning to accelerate optimization, and metric selection in proximal optimization. Before proceeding to the main contributions, related work is summarized with respect to both areas. 

\subsection{Learning to Accelerate Optimization}
Various systems for learning fast solutions to optimization problems have been proposed. For example, several works have shown how to learn heuristics such as branching rules \cite{balcan2018learning,gupta2020hybrid,khalil2016learning} and cutting planes \cite{paulus2022learning} in mixed-integer programming. An early survey \cite{bengio2020machine} provides a comprehensive summary of machine learning in combinatorial optimization. Further surveys on learning to branch \cite{lodi2017learning} and learning to cut \cite{deza2023machine} provide even more detail on the topic. To enhance the resolution of optimization problems with continuous variables, several works have also considered simplifying an optimization problem by first learning its active constraints  \cite{bertsimas2021voice,misra2022learning}.
 An altogether different paradigm aims to train deep neural networks to produce solutions to optimization problem directly. For example, several works consider end-to-end learning of solutions to combinatorial problems \cite{bello2017neural,khalil2017learning,kool2018attention,vinyals2017pointer}. Other works have shown how to learn solutions to problems with general nonlinear constraints, either by leveraging Lagrangian duality \cite{fioretto2020predicting, kotary2022fast, park2023self}, differentiable constraint corrections \cite{donti2021dc3}, or reparametrization of the feasible space \cite{konstantinov2024new}. Another closely related direction \cite{sambharya2023end} focuses on learning warm-starts to proximal algorithms for quadratic programming.

\subsection{Metric Selection in Proximal Optimization}

The potential for accelerating the convergence of a proximal algorithm by optimizing its underlying proximal metric has been theoretically demonstrated in previous works. The authors in \cite{Giselsson17} derived the optimal choice of metric for ADMM and Douglas-Rachford splitting algorithms on a limited class of problems. Based on this result, they also suggested heuristic methods for selecting an appropriate metric for problems outside of that class. Similar results were shown in \cite{giselsson2015metric} for a fast dual forward-backward splitting method.  Unfortunately, these theoretical results do not extend to many problems of practical interest, including generic Quadratic Programming (QP) problems. Furthermore, when the optimal metric can be computed, it typically requires solution of a difficult semidefinite program, reducing its practical benefit in accelerating the solution of problems. This paper demonstrates the use of end-to-end machine learning to derive models whose predicted proximal metrics can outperform the heuristic theory-based models of \cite{Giselsson17} on several QP problems.



\section{Preliminaries}

Let $\mathcal{S}^{n}_{++}$ be the set of positive definite matrices. For $M\in \mathcal{S}^{n}_{++}$ let $\mathbb{R}^{n}_{M}$ be the Hilbert space on $\mathbb{R}^{n}$ with the corresponding inner product and norm defined respectively for all $x,y \in \mathbb{R}^{n}$ as
\begin{align*}
\left< x,y \right>_{M} = x^{T}My, \qquad \text{and} \qquad
||x||_{M}^{2} = x^{T}Mx \, .
\end{align*}
We will denote by $\Gamma(\mathbb{R}^{n}_{M})$ the set of functions $f:\mathbb{R}^{n} \to \mathbb{R}\cup \{\infty\}$ that are proper, closed and convex, where $\mathbb{R}\cup \{\infty\}$ represents the extended reals. For $f,g \in \Gamma(\mathbb{R}^{n}_{M})$, \emph{Douglas-Rachford} splitting (DR) considers optimization problems of the form
\begin{equation}
\min_{x\in \mathbb{R}^{n}} f(x) + g(x) \, ,
\label{eq:gen_opt}
\end{equation}

and computes optimal solutions by following the iterations
\begin{subequations}
\label{eq:DR_alg_general}
    \begin{align}
    &y_{k} = \mbox{prox}_{\gamma}g(x_{k}) \, , \\
    &z_{k} = \mbox{prox}_{\gamma}f(2y_{k} - x_{k} ) \, , \label{eq:DR_fmin} \\
    &x_{k+1} = x_{k} + z_{k} - y_{k} \, ,
    \end{align}
\end{subequations}
where the $\mbox{prox}_{\gamma}f$ is the \emph{proximal operator} defined with respect to the space $\mathbb{R}^{n}_{M}$ as
\begin{equation}
\mbox{prox}_{\gamma}f(x) = \argmin_{z\in \mathbb{R}^{n}} f(z) + \frac{1}{\gamma}|| x - z||_{M}^{2} \, ,
\label{eq:prox_def_general}
\end{equation}
for $\gamma >0$. It can be shown that if a solution of \eqref{eq:gen_opt} exists then the DR iterations will converge, in particular the sequence $\mbox{prox}_{\gamma}g(x_{k})$ will converge weakly to a solution and under additional mild assumptions will converge strongly. Various alternative formulations and relaxations of DR exist but in this paper we will primarily restrict consideration to the formulation  \eqref{eq:DR_alg_general}; for proofs and additional details see for example \cite{Combettes2011ConvexAA}.   

This formulation naturally gives rise to the question of the selection of a positive-definite matrix $M$ to define a metric in \eqref{eq:prox_def_general} for a given problem to improve convergence of the iterations \eqref{eq:DR_alg_general}. In \cite{Giselsson17} Giselsson and Boyd study the optimal metric choice to improve convergence rate for DR applied to the Fenchel dual of \eqref{eq:gen_opt}, which can be shown to be equivalent to employing the alternating direction method of multipliers (ADMM) on the primal problem \cite{Giselsson17,Gabay1983,Ecksteinthesis}. In this case, any choice of $M$ other than the identity is equivalent to the use of preconditioning in ADMM. A standard formulation for problems to be solved by ADMM is given by
\begin{subequations}
\label{eq:gen_ADMM_problem}
\begin{align}
&\min_{x\in \mathbb{R}^{n},y\in\mathbb{R}^{m}} ~ f(x) + g(y) \\
&\mbox{subject to: }~ Ax + By =c 
\end{align}
\end{subequations}
for $f,g \in \Gamma(\mathbb{R}^{n}_{M})$, $A\in \mathbb{R}^{m,n}$, $B\in \mathbb{R}^{m,m}$, and $c\in \mathbb{R}^{m}$. ADMM applied to the preconditioned primal problem 
\begin{subequations}
\label{eq:precon_ADMM_problem}
\begin{align}
&\min_{x\in \mathbb{R}^{n},y\in\mathbb{R}^{m}} ~ f(x) + g(y) \\
&\mbox{subject to: }~ MAx + MBy =Mc \,,
\end{align}
\end{subequations}
is equivalent to DR applied to its Fenchel dual when utilizing the metric $M$. Thus dual DR using metric $M$ can be implemented by the primal ADMM iterations
\begin{subequations}
\small
\label{eq:gen_ADMM_alg}
\begin{align}
&x_{k+1} = \argmin_{x\in \mathbb{R}^{n}} \{ f(x) + \gamma 2 ||M(Ax + By_{k} - c) + u_{k}||^{2}_{2} \} \, \label{eq:ADMM_fmin},\\
&y_{k+1} = \argmin_{y}\{g(y) + \frac{\gamma}{2}||M(Ax_{k+1} + By - c) + u_{k} ||^{2}_{2} \} \, ,\\
&u_{k+1} = u_{k} + M(Ax_{k+1} + By_{k+1} - c) \, .
\end{align}
\end{subequations}
In \cite{Giselsson17} the authors show how to calculate the matrix $M$ which optimizes the convergence rate of ADMM applied to \eqref{eq:precon_ADMM_problem}, under the additional assumptions that $f$ is strongly convex and smooth, and that $A$ has full row rank. In such cases, an optimal metric $M$ can be modeled as the solution to a related semidefinite programming problem (SDP).  However, those requisite assumptions exclude many practical optimization problems, including general-form quadratic programming (QP) problems. For QP forms outside the scope of these assumptions, the authors of \cite{Giselsson17} suggest heuristic models of metric selection. 

Significant work has been done to improve ADMM convergence using preconditioning, for instance in \cite{Ghadimi2015,BENZI2002}. As pointed out in \cite{Giselsson17}, in the case of QP problems these methods ultimately amount to reconditioning the quadratic objective function. The same is true of the heuristic method they propose, which equates to selecting the optimal metric for a related \emph{unconstrained} QP. 
In this paper, we explore the potential for empirically learning metrics to enhance convergence of proximal algorithms on QP problems which do not satisfy the assumptions required for metric optimization presented in \cite{Giselsson17}. We investigate solution of QP problems using learned metrics with DR applied to both the primal and dual problems, implementing ADMM for solution of the dual as given in \eqref{eq:gen_ADMM_alg}.  

\section{Learning Metrics to Accelerate Quadratic Programming}

The proposed system for metric learning in proximal optimization leverages a reformulation of general QP problems which renders the metrics easier to learn. In this section, we first introduce the problem reformulation before describing details of the end-to-end learning approach. In brief, a neural network model is trained to predict positive definite matrices $M$ as a function of the parameters which define an optimization problem instance. Solution error after a fixed number of iterations \eqref{eq:DR_alg_general} or \eqref{eq:gen_ADMM_alg}  is treated as a loss function and minimized, by backpropagation through the solver iterations in stochastic gradient descent training. While the system is general and can in principle be applied to any problem of the form \eqref{eq:gen_opt}, the scope of this paper is limited to demonstration on QP problems.

\subsection{Problem Reformulation}
Let $Q\in \mathbb{R}^{n,n}$ be a positive semi-definite matrix, $q\in \mathbb{R}^{n}$, $L \in \mathbb{R}^{m,n}$, $b\in \mathbb{R}^{m}$, $W\in \mathbb{R}^{k,k}$, and $c \in \mathbb{R}^{k}$. We consider QP problems of the form
\begin{subequations}
\label{eq:QP_std}
\begin{align}
    \argmin_{x \in \mathbb{R}^{n}} &\;\;
    \frac{1}{2} x^T Q x + q^T x    \\
    \textit{s.t.} \;\;
    &       Lx = b \\
    &       Wx + c \leq 0. \label{QP_std_ineq}
\end{align}
\end{subequations}
For implementation with both primal DR and ADMM we introduce slack variables $s\in \mathbb{R}^{m}$ and  reformulate the problem as 
\begin{subequations}
\label{eq:QP_std_slacks}
\begin{align}
    \argmin_{z \in \mathbb{R}^{n+k}} &\;\;
    \frac{1}{2} z^T I_{x}^TQI_{x} z + q^TI_{x} z     \\
    \textit{s.t.} \;\;
    &       Rz + r = 0  \label{eq:gen_constraint_set}\\
    &       I_{s}z \geq 0,
\end{align}
\end{subequations}
where  $I^n$ is the $n\times n$ identity matrix, and

\begin{align*}
&z = \begin{bmatrix}x \\ s \end{bmatrix} \, , \quad R = \begin{bmatrix} L & 0 \\ W &  I^k \end{bmatrix} \, , \quad r = \begin{bmatrix} -b \\ c \end{bmatrix}
\, , \\
&\quad I_{s} = \begin{bmatrix} 0 & 0 \\ 0 & I^k \end{bmatrix} \, , \quad I_{x} = \begin{bmatrix} I^n & 0 \\ 0 & 0 \end{bmatrix} \, .
\end{align*}

Note that even if the inequalities \eqref{QP_std_ineq} represent simple box constraints on the variables $x$ (for example $x<0$) we still introduce corresponding slack variables. This is done to construct a splitting for both primal DR and ADMM with the intent to increase the impact $M$ can have on convergence, as will be described in the next section. 

\subsubsection{QP Splitting}
For implementation of both primal DR and ADMM on problem \eqref{eq:QP_std_slacks} we use the splitting
\begin{subequations}
\label{eq:QP_DR_splitting}
\begin{align}
    &f(z) = z^T I_{x}^TQI_{x} z + q^TI_{x} z + i_{\{ z\in \mathbb{R}^{n+k}~:~Rz + r = 0\} }(z) \\
    &g(z) = i_{\{ z\in \mathbb{R}^{n+k}~:~ I_{s}z \geq 0\} }(z)
\end{align}
\end{subequations}
where for a set $\mathcal{S}$ we define the indicator function on $\mathcal{S}$ to be
\begin{equation*}
i_{S}(x) = \begin{cases} 0 & x \in S \\ \infty & x \notin S \;\;. \end{cases}
\end{equation*}
With this splitting we implement the primal DR iteration as given in \eqref{eq:DR_alg_general}, and implement ADMM as in \eqref{eq:gen_ADMM_alg}, with $A = I$, $B = -I$, and $c = 0$. Minimization steps with respect to $f$ can be accomplished with for example the corresponding Karush-Kuhn-Tucker conditions. Minimization steps with respect to $g$ for both algorithms equate to projections onto the positive orthant. Slack variables are initialized at zero throughout.

For both ADMM and DR iterations using the splitting \eqref{eq:QP_DR_splitting}, the proximal operators as given in \eqref{eq:DR_fmin} and \eqref{eq:ADMM_fmin} equate to projections onto the relaxed constraint set \eqref{eq:gen_constraint_set} with respect to the underlying metric. With the slack variables  for each inequality constraint initialized at zero, a relatively large corresponding weight in the metric matrix $M$ will bias the non-Euclidean projection to maintain those slacks near zero. Hence if $M$ has relatively large weights for just the slacks corresponding to the active constraints of a problem, the projection can approximate projection onto the active set. Indeed, the learned metrics exhibit this expected behavior as illustrated in Figure \ref{fig:optimal solution and metric selection visualization}.

\subsection{End-to-End Learning Framework}
\label{sec:parameteric optimization}

As suggested by the results of Section \eqref{sec:NumericalResults}, the optimal metric for solving of an instance of \eqref{eq:QP_std_slacks} can be closely related to the active constraints at its optimal solution. Thus, it is expected that learning the optimal metrics for solving a class of problems may be nearly as difficult as learning their optimal solutions. As is common in prior works on learning to solve optimization problems, the metric learning problem is formulated relative to a \emph{parametric} optimization problem

\begin{equation}
x^{\star}(p) = \argmin_{x\in \mathbb{R}^{n}} f_p(x) + g_p(x) \, ,
\label{eq:gen_opt_parametric}
\end{equation}

and we learn to predict metrics for parametric problem instances within a limited distribution.  In the QP problem \eqref{eq:QP_std}, this corresponds to the elements $Q$, $q$, $L$, $b$, $w$, and $c$ each being potential functions of $p$. The metric $M$ which best solves problem \eqref{eq:gen_opt_parametric} is then learned as a function of the problem's parameters $p \in \mathbb{R}^v$. This learned function takes the form of a neural network  $\mathcal{N}_{\omega} : \mathbb{R}^{v} \to \mathcal{S}^{n}_{++}$ with weights $\omega$, so that $M = \mathcal{N}_{\omega}(p)$. It is trained over a distribution of problem parameters $p \sim \mathcal{P}$, for which a finite dataset of instances $\{p_{i}\}_{i\in \mathcal{T}}$ are drawn. A target dataset $\{x^{*}(p_{i})\}_{i \in \mathcal{T}}$  contains the corresponding optimal solutions, as per \eqref{eq:gen_opt_parametric}.

To define a loss function for training $\mathcal{N}_{\omega}$ on these data, first define the following function. Let $\mathcal{D}_k:\mathbb{R}^{v}\times \mathcal{S}^{n}_{++} \times \mathbb{R}^{n} \to \mathbb{R}^{n}$ denote the application of $k$ iterations of DR or ADMM, on problem \eqref{eq:gen_opt_parametric} with parameters $p$ using metric $M$, starting from initial variable values $x_0$. It yields a solution estimate $x_k$; that is, $\mathcal{D}_k(p,M,x_0) = x_k$. The metric prediction model $\mathcal{N}_{\omega}$ is then trained to minimize the overall loss function
\begin{equation}
\label{loss_function}
\min_{\omega} \frac{1}{|\mathcal{T}|} \sum_{i \in \mathcal{T}} || \mathcal{D}_k(p_{i}, \mathcal{N}_{\omega}(p_i),\mathcal{E}_{\theta}(p_i) )  - x^{*}(p_{i})||^{2}
\end{equation}
by stochastic gradient descent. This requires backpropagation of gradients through the solver iterations which constitute $\mathcal{D}_k$. In this work, backpropogation is performed by automatic differentiation in PyTorch \cite{paszke2017automatic}.

In equation \eqref{loss_function}, the function $\mathcal{E}_{\theta}$ is an oracle which returns a starting point $x_0$ for any parameter vector $p$. As part of an overall mechanism for producing fast solutions to \eqref{eq:QP_std}, it is a neural network trained to produce direct estimates of the optimal solution to \eqref{eq:gen_opt_parametric} by mean square error regression:

\begin{equation}
\min_{\theta} \frac{1}{|\mathcal{T}|} \sum_{i \in \mathcal{T}} || \mathcal{E}_{\theta}(p_i) - x^{*}(p_{i})||^{2} \, .
\end{equation}

\subsection{Metric Representation}
Finally, we describe the manner of representation used to predict the metrics $M$ via the model $M = \mathcal{N}_{\omega}(p)$. As a neural network, $\mathcal{N}_{\omega}$ produces a vector of values $m \in \mathbb{R}^n$ which is then scaled between predefined upper and lower bounds $[m_{\min}, m_{\max}]$. Finally, a scalar parameter $\rho \in \mathbb{R}$ is predicted and also scaled to fit within predefined bounds $[\rho_{\min},\rho_{\max}]$. The final metric is constructed as the diagonal matrix $M = \textit{diag}(\rho \cdot m)$.

\section{ Numerical Results}
\label{sec:NumericalResults}

For illustrative purposes, this section begins with a reductive two-dimensional problem on which some effects of metric learning are most easily observed. Then, we demonstrate the effect of metric learning on convergence for larger examples consisting of a portfolio optimization problem, and a model predictive control problem.

In the following experiments, predictive models $\mathcal{N}_{\omega}$ and $\mathcal{E}_{\theta}$ are fully connected neural networks with rectified linear unit (ReLU) activation functions. The values $\rho_{\min}, \rho_{\max}, m_{\min}, m_{\max}$ can be treated as hyperparameters; in practice, it is found that effective metrics can be learned by searching over $\rho_{\max}$ while the others remain fixed. All numerical test cases in this work are implemented using NeuroMANCER, an open source differentiable programming library built on top of Pytorch \cite{Neuromancer2023}.

\subsection{Active Set Prediction}
\begin{figure}[htbp]
     \centering
     \begin{subfigure}[b]{0.35\textwidth}
         \centering
         \includegraphics[width=\textwidth]{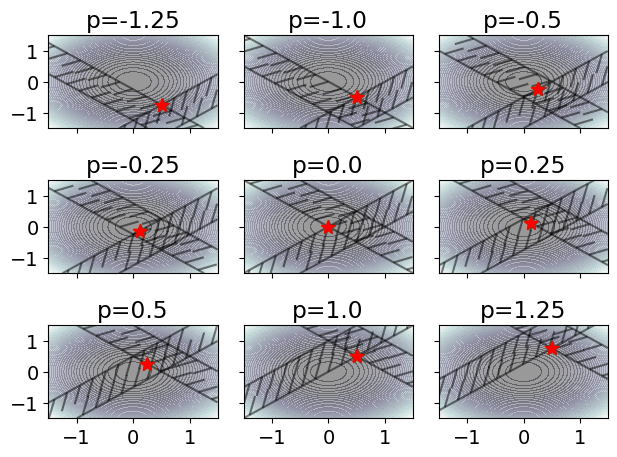}
         \caption{Optimal Solution}
         \label{fig:opt_loc}
     \end{subfigure}
     \hfill
     \begin{subfigure}[b]{0.35\textwidth}
         \centering
         \includegraphics[width=\textwidth]{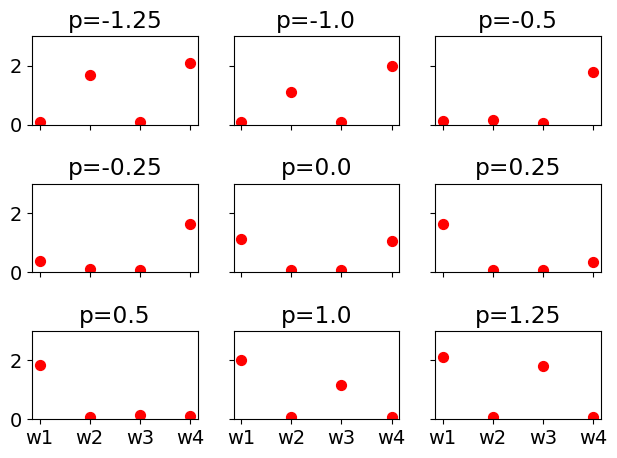}
         \caption{Metric weight for each slack variable}
         \label{fig:metric_weights}
     \end{subfigure}
        \caption{Plot (a) shows the optimal solution of \eqref{eq:pqp_problem} for parameter choice $p_{1} = p_{2} = p$, for values of $p$ ranging from $-1.25$ to $1.25$. As the feasible set is translated around the origin the optimal solution marked by a red star can be seen to trace along the constraints. Correspondingly, plot (b) shows that the learned DR metric weights corresponding to the slack variables for the inequality constraints are near zero when the constraint is not active and larger when the constraint is active, showing a clear correspondence between the active set and metric weights on slack variables.}
        \label{fig:optimal solution and metric selection visualization}
\end{figure}

This section illustrates how metric learning correlates with active set prediction in inequality-constrained problems, by assigning higher metric weights to the coordinates which correspond to slack variables on the problem's active constraints. As an illustrative example, we consider a simple QP problem:

\begin{subequations}
\label{eq:pqp_problem}
\begin{align}
&\min_{x,y} ~ x^{2} + y^{2} \\
&\mbox{subject to:} \nonumber\\
& -x -y +p_{1} \leq 0 \\
& x + y - p_{1} -1 \leq 0 \\
& x - y + p_{2} - 1 \leq 0 \\
& -x + y - p_{2} \leq 0
\end{align}
\end{subequations}
for parameters $p_{1},p_{2} \in [-2,2]$. The constraint set defines a box which is translated around the origin according to the parameter choices as shown in Figure \ref{fig:optimal solution and metric selection visualization}. After assigning slack variables $s$, the constraints become
\begin{subequations}
\label{eq:pqp_slack_constraints}
\begin{align}
& -x -y +p_{1} + s_{1} = 0 \\
& x + y - p_{1} -1 + s_{2} = 0 \\
& x - y + p_{2} - 1 + s_{3}  = 0 \\
& -x + y - p_{2} + s_{4} = 0 \\
& s_{1}, ~ s_{2}, ~ s_{3}, ~ s_{4}~ \geq 0
\end{align}
\end{subequations}

We train both ADMM and DR metrics over the parameter space. The neural network map $\mathcal{N}_{\omega}$ returns diagonal metrics with diagonals of the form
\[ \textit{diag}([w_{y},w_{x},w_{1},w_{2},w_{3},w_{4}])
\]
where the weights $w_{x}$ and $w_{y}$ correspond to the primal variables, and the weights $\{w_{1},w_{2},w_{3},w_{4}\}$ correspond to the slack variables for each of the constraints. 

\paragraph{Settings} The initial prediction model $\mathcal{E}_{\theta}$ uses hidden dimension 80. For $\mathcal{N}_{\omega}$ we use hidden dimension 20, with bounds $\rho_{\min}=0.05$, $\rho_{\max}=1.0$, $m_{\max} = 5.0$, and $m_{\min}=0.2$. We sample 2000 parameters uniformly at random to construct a training set, and an additional 2000 parameters uniformly at random for a test set. The initial solution estimator $\mathcal{E}_{\theta}$ was trained for 200 epochs at learning rate of $0.001$. Both ADMM and DR were run for 10 steps during training, and $\mathcal{N}_{\omega}$ was trained for 100 epochs at a learning rate of $0.001$ for both.

\begin{figure}[htbp]
    \centering
    \includegraphics[width=0.8\linewidth]{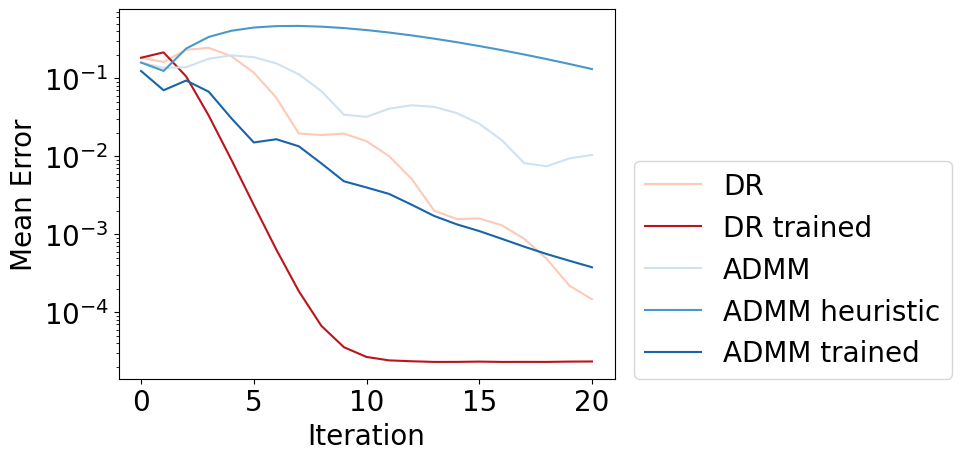}

    \caption{Comparison of convergence of DR and ADMM using trained metrics versus not, as well as comparison to use of a heuristic metric choice. Reported values are the mean error at each iteration on 2000 test set problems.}
    \label{fig:pqp_convergence}
\end{figure}

\paragraph{Results} As shown in Figure \ref{fig:pqp_convergence}, DR with a trained metric converges the fastest, while ADMM using the heuristic metric as presented in \cite{Giselsson17} converges most slowly. Computation of the heuristic metric effectively ignores the inequality constraints, and computes a metric that achieves the fastest convergence with respect to the objective, were no constraints present this metric would achieve the best possible convergence. However, the results show that in the presence of constraints it can be detrimental.

\begin{figure}[htbp]
     \centering
     \begin{subfigure}[h]{0.21\textwidth}
         \includegraphics[width=\linewidth]{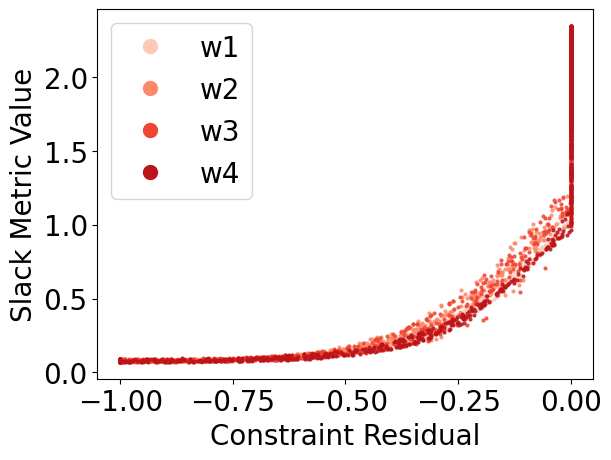}
         \caption{DR}
         \label{fig:scatter_one}
     \end{subfigure}
     \hspace*{\fill}
     \begin{subfigure}[h]{0.21\textwidth}
         \includegraphics[width=\linewidth]{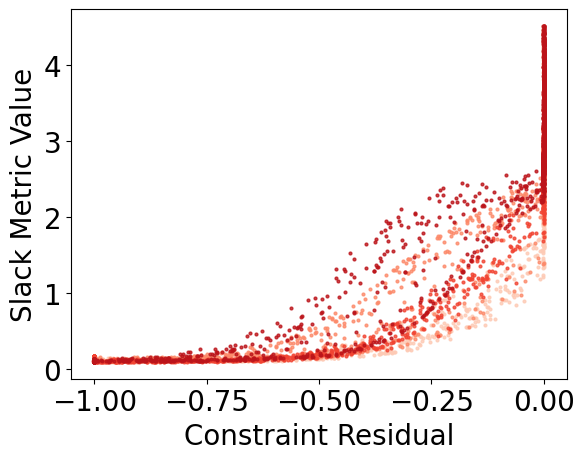}
         \caption{ADMM}
         \label{fig:scatter_two}
     \end{subfigure}
        \caption{ Plots of the learned metric weight for slack variables and the corresponding constraint residual at the optimum on two thousand test problems for metrics learned using DR and ADMM. As constraint residuals becomes zero the constraints become active and the metric weight for the corresponding slack increases.}
        \label{fig:metric weight constraint relation}
\end{figure}

Conversely, the learned metrics appear to achieve faster convergence by incorporating information about the active constraints. Figure \ref{fig:optimal solution and metric selection visualization} and Figure \ref{fig:metric weight constraint relation} highlight the close correspondence between the metric weights corresponding to slack variables and whether the associated constraints are active at the optimum for a problem.

\subsection{Portfolio Optimization Problem}
\label{subsec:portfolio}

This experiment models the optimal allocation of assets in an investment portfolio as a quadratic programming problem. Given $n$ investment assets, their future price differentials $p \in \mathbb{R}^n$ are treated as parameters in the following QP:
\begin{subequations}
\label{eq:portfolio_problem}
\begin{align}
x^{*}(p) = \argmin_{x} ~ & x^{T} \Sigma x - p^{T}x   \label{eq:portfolio_problem_objective} \\
\mbox{subject to:} \;\;\;
& 1^{T}x = 1 \label{eq:portfolio_problem_eq}  \\
& \;\;\;\; x \geq 0, \label{eq:portfolio_problem_ineq}
\end{align}
\end{subequations}
where $\Sigma$ represents a constant covariance matrix. The objective \eqref{eq:portfolio_problem_objective} balances maximization of future profit with minimization of price covariance as a measure of risk. Constraints (\ref{eq:portfolio_problem_eq},\ref{eq:portfolio_problem_ineq}) define a valid proportional allocation.

\paragraph{Settings}
Data on price action per asset are collected from the Nasdaq online database \cite{NASDAQ}. A training dataset of $5000$ observations for the future price differential $p$ are generated by adding Gaussian random noise to this data, plus an additional $500$ each for the validation and testing sets. A $5$-layer neural network with hidden layer size $n$ is used to predict the elements of a diagonal metric matrix $M$ as a function of $p$. The following parameters are fixed: $m_{\min} = 0.01$, $m_{\max} = 1.0$, $\rho_{\min} = 0.01$. A search over the upper bound $\rho_{\max} \in \{ 1.0, 5.0, 10.0, 50.0, 100.0, 500.0 \}$ shows that the best results occur for $\rho_{\max} = 100.0$ but remain similar for higher values of $\rho_{\max}$.

\begin{figure}[htbp]
    \centering
    \includegraphics[width=1.0\linewidth]{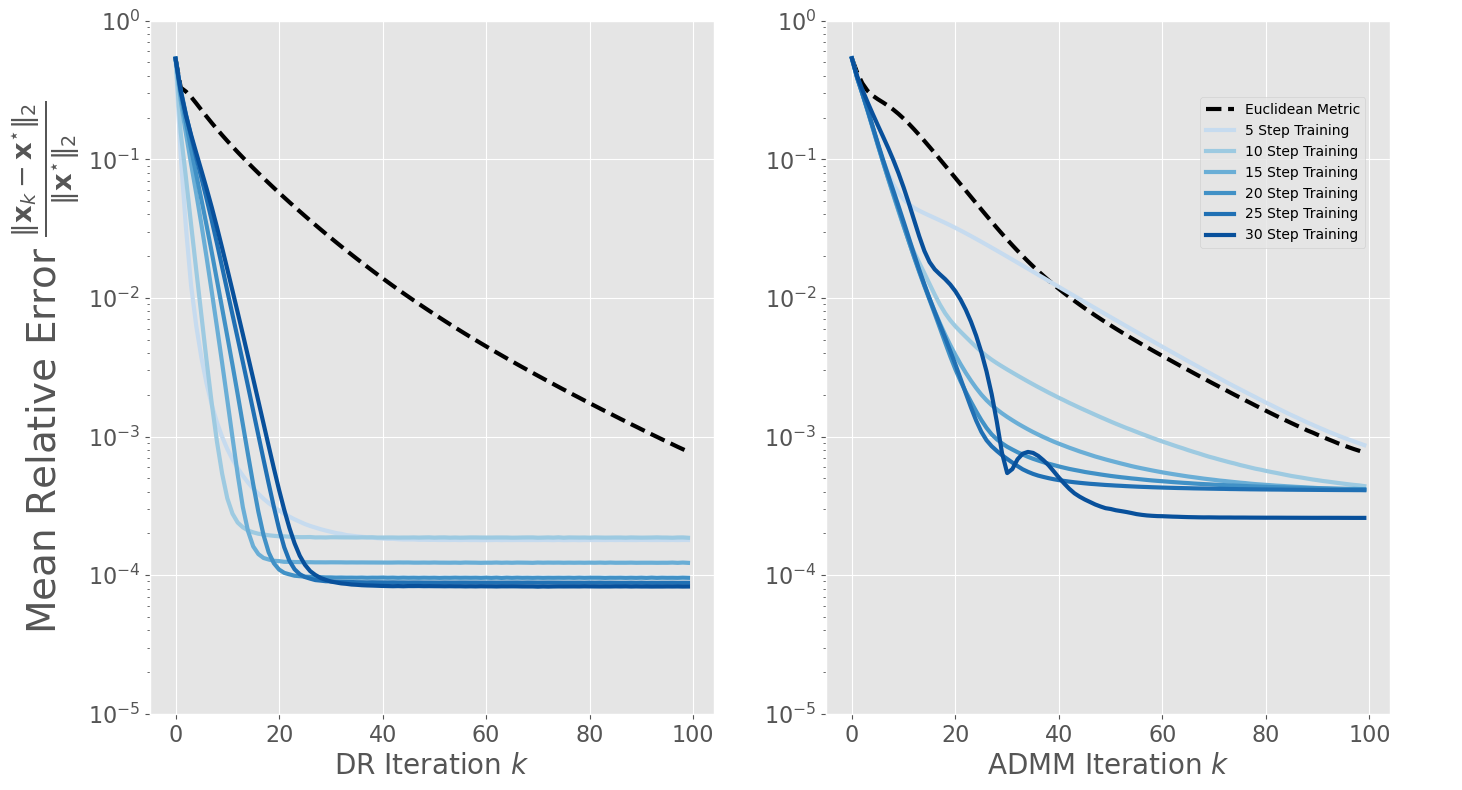}

    \caption{Results of training proximal metrics for DR and ADMM on Portfolio Optimization, to minimize error at increments of $5$ iterations. }
    \label{fig:portfolio_convergence}
\end{figure}

\paragraph{Results}
Figure \ref{fig:portfolio_convergence} illustrates convergence of the DR and ADMM algorithms in solving \eqref{eq:portfolio_problem}, under metric prediction trained with $k$ iterations in the loop, for $k \in \{5,10,15,20,25,30\}$. At test time, $100$ iterations of DR and $150$ iterations ADMM are applied regardless of $k$. The plotted values represent mean relative solution error in the $L2$ norm. Dotted black curves correspond to a baseline in which the standard Euclidean metric is used. 

The following observations apply to both DR (at left) and ADMM (at right). The metric prediction model which is trained for $k$ iterations always attains the best accuracy at exactly $k$ iterations. Additionally, the models trained with more iterations $k$ perform better as more iterations are performed at test time. This implies that an accelerated DR or ADMM model intended for exactly $k$ iterations should be trained using $k$ iterations, while a model intended for iteration until convergence should train using large values of $k$.

Note additionally that in this particular experiment, training for small $k$ comes at a cost of long-term convergence in the case of DR. In ADMM, models trained with larger $k$ generally perform equally or better than with smaller $k$. An exception is observed for $k=30$ in ADMM, in which error is minimized at iteration $30$ at the cost of higher error in both earlier and later iterations. In all but one case, the learned metrics outperform the Euclidean metric at test time.

\paragraph{Comparison with Heuristic Metric Selection}
In order to compare the proposed metric learning for ADMM against the theoretically prescribed heuristic metric choice given in \cite{Giselsson17} we test the metric learning experiment on problem \eqref{eq:portfolio_problem} with a reduced size of $n=20$ assets. Problem size is reduced to allow a heuristic metric to be calculated efficiently.

\begin{figure}[htbp]
    \centering
    \includegraphics[width=1.0\linewidth]{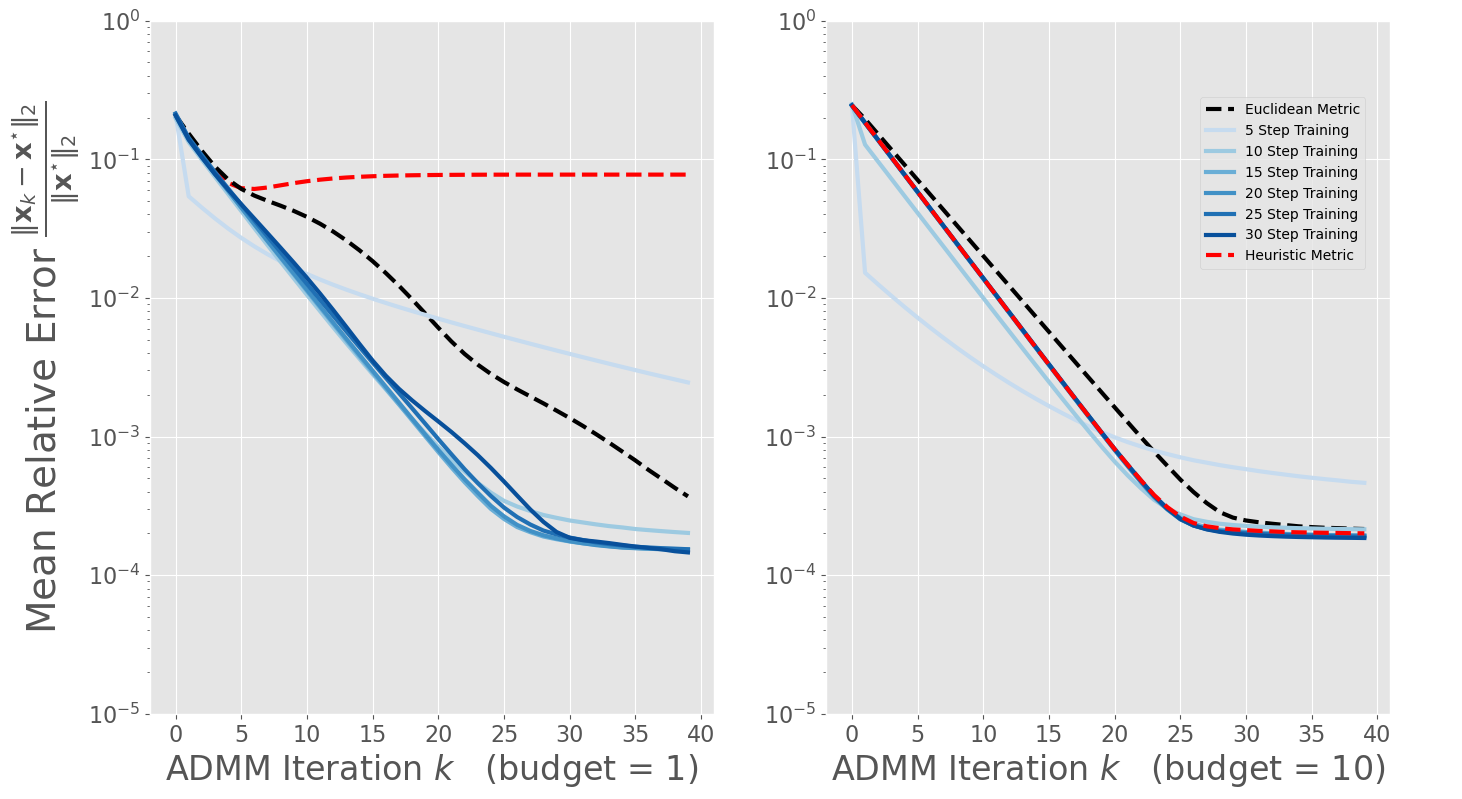}
    \caption{Comparison of trained and heuristic metrics for ADMM for Portfolio Optimization. The left plot presents a case where problem constraints are routinely active over problem parameters, while the right shows a case in which problem constraints are more rarely active.}
    \label{fig:portfolio_convergence_withHeur}
\end{figure}

To illustrate the effect of active constraints on the viability of the heuristic metric, results for ADMM are compared on two variants of problem \eqref{eq:portfolio_problem}. The first is as given in \eqref{eq:portfolio_problem}, and the second replaces the equality constraint \eqref{eq:portfolio_problem_eq} with a scaled asset allocation budget $ 1^{T}x = 10 $. This increase in budget has the effect of reducing the percentage of active constraints over problem parameters. Figure \ref{fig:portfolio_convergence_withHeur} shows the results due to each allocation budget, at left and right respectively. In the larger budget case few constraints are active on average over the test set. This approximates a problem setting with no constraints where the heuristic metric would be optimal, and it can be seen to improve convergence in comparison to the Euclidean metric. Further, the learned optimal metrics perform similarly when trained for more than $k=5$ iterations. On the other hand, the smaller budget leads to more active constraints and the heuristic metric performs poorly. Conversely, the trained metric achieves a greater improvement in convergence rate with respect to the Euclidean metric than in the high budget case, highlighting the capacity for improvement by incorporating information about active constraints.

\subsection{Quadcopter Control}
We also test metric learning on a standard reference tracking model predictive control problem implemented for a linear discrete time dynamical model of a quadcopter. Let the dynamics be defined by $A\in \mathbb{R}^{n,n}$, $B\in \mathbb{R}^{n,m}$, and the cost function be defined by positive semi-definite matrices $Q\in \mathbb{R}^{n,n}$ and $R \in \mathbb{R}^{m,m}$. Here we take the parameter $p\in \mathbb{R}^{n}$ to be the initial state of the system and solve
\begin{subequations}
\small
\begin{align}
u^{*}(p), x^{*}(p) = &\argmin_{x,u} \sum_{k = 0}^{N} (x_{k+1} - r)^{T}Q(x_{k+1} - r) + u_{k}^{T}Ru_{k} \\
&\mbox{subject to: } ~ \nonumber \\
&x_{1} = Ap + Bu_{0} \, , \\
&x_{k+1} = Ax_{k} + Bu_{k} \, , \\
&u_{a}\leq u_{k} \leq u_{b} \, , \\ 
&x_{a} \leq x_{k} \leq x_{b}\, , \\
&\forall k \in \{1,2,\dots N\} \, . 
\end{align}
\label{eq:quadQP}
\end{subequations}
The optimal solution produces control actions $u_{k} \in \mathbb{R}^{m}$ that drive the state of the system $x_{k} \in \mathbb{R}^{n}$ from a given initial state $p$ towards the reference point $r \in \mathbb{R}^{n}$ over a finite number of time steps $ k \in \{1,2,\dots N\}$. The control actions must also keep the state within the bounds $x_{a},x_{b} \in \mathbb{R}^{n}$ while staying within the control bounds $u_{a},u_{b} \in \mathbb{R}^{m}$. In practice the problem \eqref{eq:quadQP} is solved iteratively with the control implemented for just the initial time step, then the problem is re-solved from the new system state at the next time step. Thus the problem can be understood to be parameterized by the initial state and reference point. 

Here we assume a fixed reference point at the origin and sample over a range of initial system states. The matrices describing the quadcopter model dynamics and objectives as well as state and control bounds are given in the Appendix \ref{apx:quadcopter}, they were taken from a set of benchmark problems curated in \cite{Ferreau2015}, with data from the repository \cite{FerreauGithub}.

\paragraph{Settings}
The quadcopter model has a 12 dimensional state and 
4 dimensional control input. Only the first two state variable are constrained, and are restricted to the interval $[-\pi / 6 ~,~ \pi/6]$. To generate training data we take the reference point $r$ to be the origin and we generate initial states $p$ uniformly at random with the first two variables sampled from $[-\pi / 6 ~,~ \pi/6]$ and the rest from $[-0.8,0.8]$. This choice was made such that generated problems were feasible, and state constraints were routinely active at solutions. We solve the problem \eqref{eq:quadQP} over a 10 step horizon, resulting in $n = 304$ variables with 132 inequality constraints, and 132 equality constraints. Diagonal elements of a metric matrix $M$ are predicted on a per-instance basis using a $5$-layer ReLU network of hidden layer size $400$. As in Section \ref{subsec:portfolio}, the parameter choices: $m_{\min} = 0.01$, $m_{\max} = 1.0$, $\rho_{\min} = 0.01$ are fixed. A search over the upper bound $\rho_{\max} \in \{ 1.0, 5.0, 10.0, 50.0, 100.0, 500.0 \}$ shows that the best results occur for $\rho_{\max} = 50.0$ and remain similar for higher values of $\rho_{\max}$.

\paragraph{Results}

\begin{figure}[htbp]
    \centering
    \includegraphics[width=1.0\linewidth]{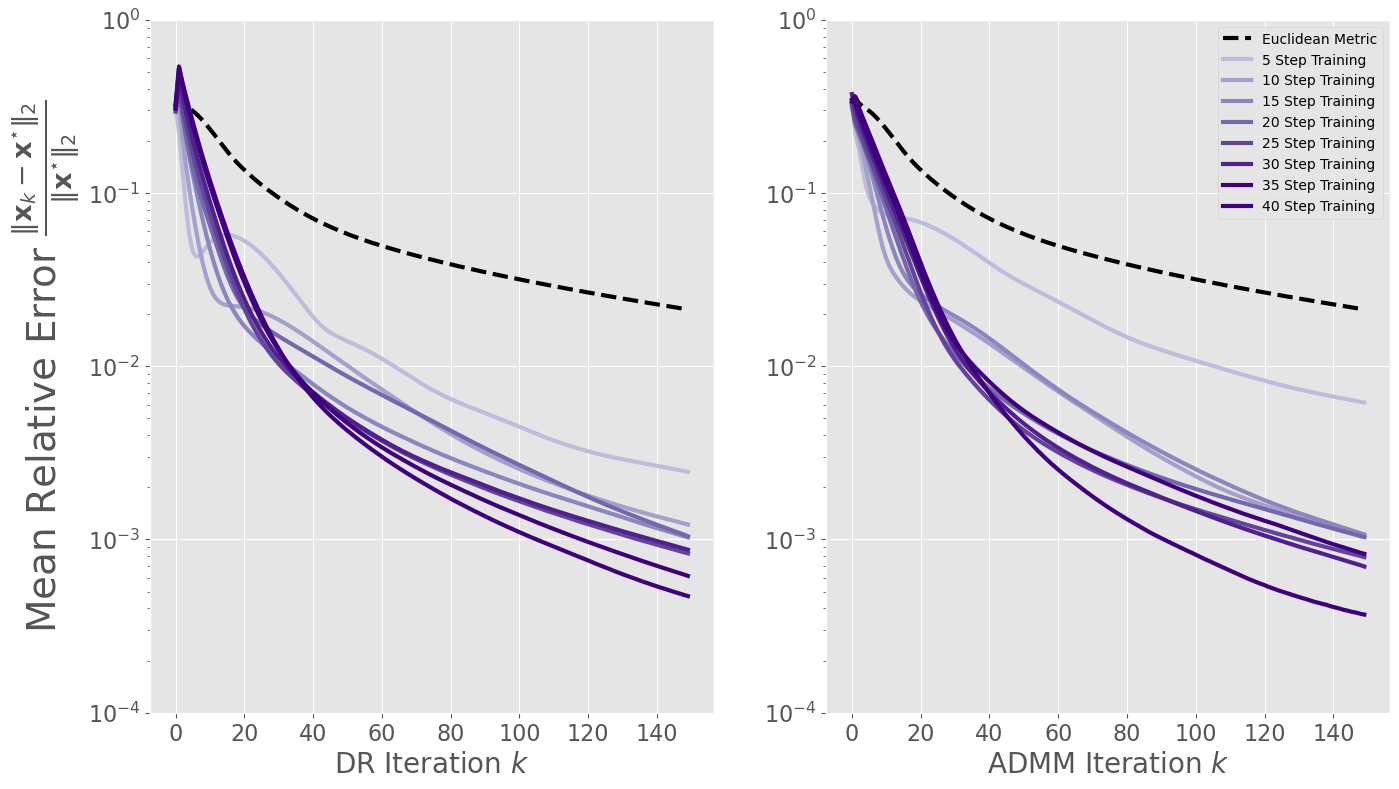}
    \caption{Results of training proximal metrics for DR and ADMM on Quadcopter Control, to minimize error at increments of $5$ iterations. }
    \label{fig:quadcopter_convergence}
\end{figure}

Convergence of both DR and ADMM due to the various trained metric prediction models are illustrated in Figure \ref{fig:quadcopter_convergence}. Prediction models are trained using $k$ steps of each algorithm for $\{5,10,15,20,25,30,35,40\}$. Solution error over the full horizon needed for convergence is not shown, to make visible the effects of training up to the first $40$ iterations. This is consistent with the intended application in real-time optimization, which demands solutions within stringent time constraints. 

With regards to the effect of $k$, similar observations apply as in the portfolio optimization experiments. Models trained to minimize error at iteration $k$ consistently perform best after exactly $k$ iterations. Meanwhile, training with larger $k$ generally benefits long-term convergence. Note that nearly all trained models reach a relative error of $1e-2$ in a fraction of the iterations required by the standard variant with a Euclidean metric.

\section{CONCLUSION}

Metric learning as presented here for parametric QP problems can consistently result in orders of magnitude improvements in solution accuracy at low a number of iterations. The most benefit is observed in problem settings with a significant number of inequality constraints relative to the problem size that are routinely active over parameters of interest. Notably, this is exactly the case in which the theoretically prescribed heuristic metrics are not guaranteed to be optimal.  
Future work is needed to understand the capacity for metric learning to reduce solution time on large-scale problems. By relying a problem reformulation with slack variables, the total number of variables is expanded resulting in a larger overall problem size. On the other hand, it allows for metric learning to potentially significantly reduce the number of iterations required to achieve a given accuracy as seen here. Because it is independent of other strategies for learning to accelerate optimization, it may have significant potential to be combined with previously proposed techniques. Combining the proposed metric learning with nonoverlapping strategies such as  prediction of solution warmstarts  may further reduce overall solution times. 




\section*{APPENDIX}

\subsection{Quadcopter Model Details}
\label{apx:quadcopter}

    
\begin{subequations}
    
\begin{align*}
\small
&Q = \textit{diag}([0,0,10,10,10,10,0,0,0,5,5,5], \\
&R = \textit{diag}([0.1,0.1,0.1,0.1]),\\
&u_{a} = [9.6, 9.6, 9.6, 9.6] ~ - ~10.5916, \\
&u_{b} = \;[13., 13., 13., 13]~ - ~10.5916 \\
\end{align*}
\end{subequations}
\begin{equation*}
\small
A = \begin{psmallmatrix} 1.0 & 0& 0& 0 &0 &0 &0.1& 0 &0 &0 &0 &0\\
0 &1.0 & 0 &0 &0 &0 &0 &0.1& 0 &0 &0 &0\\
0 &0 &1.0 & 0 &0 &0 &0 &0 &0.1& 0 &0 &0\\
0.0488& 0 &0 &1.0 & 0 &0 &0.0016& 0 &0 &0.0992& 0 &0\\
0 &-0.0488& 0 &0 &1.0 & 0 &0 &-0.0016& 0 &0 &0.0992& 0\\
0 &0 &0 &0 &0 &1.0 & 0 &0 &0 &0 &0 &0.0992 \\
0 &0 &0 &0 &0 &0 &1.0 & 0 &0 &0 &0 &0\\
0 &0 &0 &0 &0 &0 &0 &1.0 & 0 &0 &0 &0\\
0 &0 &0 &0 &0 &0 &0 &0 &1.0 & 0 &0 &0\\
0.9734& 0 &0 &0 &0 &0 &0.0488& 0 &0 &0.9846& 0 &0\\
0 &-0.9734& 0 &0 &0 &0 &0 &-0.0488& 0 &0 &0.9846& 0\\
0 &0 &0 &0 &0 &0 &0 &0 &0 &0 &0 &0.9846
\end{psmallmatrix} 
\end{equation*}

\begin{equation*}
\small
B = \begin{psmallmatrix} 0 &-0.0726& 0 &0.0726 \\
-0.0726& 0 &0.0726& 0\\
-0.0152& 0.0152& -0.0152& 0.0152 \\
-0 &-0.0006& -0 &0.0006 \\
0.0006& 0 &-0.0006& 0.0000 \\
0.0106& 0.0106& 0.0106& 0.0106 \\
0& -1.4512& 0 &1.4512 \\
-1.4512& 0 &1.4512& 0\\
-0.3049& 0.3049& -0.3049& 0.3049 \\
-0 &-0.0236& 0 &0.0236 \\
0.0236& 0 &-0.0236& 0\\
0.2107& 0.2107& 0.2107& 0.2107
\end{psmallmatrix} \, .
\end{equation*}

\section*{ACKNOWLEDGMENT}

This research was supported by 
the AT Scale and Data Model Convergence (DMC) initiatives
via the Laboratory Directed Research and Development (LDRD) investments at Pacific Northwest National Laboratory (PNNL). PNNL is a national laboratory
operated for the U.S. Department of Energy (DOE) by Battelle Memorial Institute under Contract
No. DE-AC05-76RL0-1830.




\newpage

\bibliographystyle{IEEEtranS.bst}
\bibliography{bib.bib}

\end{document}